\documentclass{article}
\usepackage[utf8]{inputenc}

\usepackage{amssymb}
\usepackage{amsmath}
\usepackage{amsthm}
\usepackage{graphicx}
\usepackage{booktabs}
\usepackage{siunitx}
\usepackage{hyperref}
\usepackage{algorithm}
\usepackage{algpseudocode}
\usepackage{xspace}
\usepackage{xcolor}
\usepackage{subfigure}
\usepackage{relsize}

\newcommand{\argmin}{\operatorname*{argmin}}

\newcommand{\bw}{\mathbf{w}}
\newcommand{\bg}{\mathbf{g}}
\newcommand{\bh}{\mathbf{h}}

\newcommand{\bpi}{\boldsymbol{\pi}}
\newcommand{\bdelta}{\boldsymbol{\delta}}

\newcommand{\grad}[1]{\nabla_{#1}}
\newcommand{\gradgrad}[1]{\nabla^2_{#1}}

\newcommand{\ttheta}{\tilde{\theta}}
\newcommand{\bttheta}{\boldsymbol{\ttheta}}

\newcommand{\gradgradttheta}{\nabla^2_{\ttheta} }
\newcommand{\gradbttheta}{\nabla_{\bttheta} }

\newcommand{\hy}{\hat{y}}

\newcommand{\hyu}[1]{\hat{y}(u,#1)}

\newcommand{\ICD}{iCD\xspace}
\newcommand{\IALS}{iALS\xspace}
\newcommand{\IALSpp}{iALS\nolinebreak[4]\hspace{-.05em}\protect\raisebox{.4ex}{\smaller[3]\textbf{++}}\xspace}

\newcommand{\sourcecodelink}{https://github.com/google-research/google-research/tree/master/ials/}
\newcommand{\citetuning}{\cite{rendle:ials_tuning}}

\renewcommand{\O}{\mathcal{O}}

\newcommand{\myfigurewidth}{\textwidth}

\title{\IALSpp: Speeding up Matrix Factorization\\with Subspace Optimization}

\author{
  \and
  Steffen Rendle\thanks{Google Research, Mountain View, USA}\\
  \texttt{srendle@google.com}
  \and
  Walid Krichene\footnotemark[1]\\
  \texttt{walidk@google.com}
  \and
  Li Zhang\footnotemark[1]\\
  \texttt{liqzhang@google.com}
  \and
  Yehuda Koren\thanks{Google, Haifa, Israel}\\
  \texttt{yehuda@google.com}
}
\date{\vspace{-6.0ex}}

\begin{document}

\maketitle

\begin{abstract}
\IALS is a popular algorithm for learning matrix factorization models from implicit feedback with alternating least squares.
This algorithm was invented over a decade ago but still shows competitive quality compared to recent approaches like VAE, EASE, SLIM, or NCF.
Due to a computational trick that avoids negative sampling, \IALS is very efficient especially for large item catalogues.
However, \IALS does not scale well with large embedding dimensions, $d$, due to its cubic runtime dependency on $d$.
Coordinate descent variations, \ICD, have been proposed to lower the complexity  to quadratic in $d$.
In this work, we show that \ICD approaches are not well suited for modern processors and can be an order of magnitude slower than a careful \IALS implementation for small to mid scale embedding sizes ($d \approx 100$) and only perform better than \IALS on large embeddings $d\approx 1000$.
We propose a new solver \IALSpp that combines the advantages of \IALS in terms of vector processing with a low computational complexity as in \ICD.
\IALSpp is an order of magnitude faster than \ICD both for small and large embedding dimensions.
It can solve benchmark problems like Movielens 20M or Million Song Dataset even for 1000 dimensional embedding vectors in a few minutes.
\end{abstract}

\section{Introduction}

iALS~\cite{hu:ials}, invented over a decade ago, is a popular algorithm for learning matrix factorization models from implicit feedback with alternating least squares.
The algorithm runs in linear time in the input data size and hence is broadly adopted on large datasets for its excellent scalability.
It also shows competitive quality~\citetuning{} compared to recent and often much more expensive approaches like variational auto encoders (VAE)~\cite{liang:vae}, EASE~\cite{steck:ease}, SLIM~\cite{ning:slim}, or neural collaborative filtering (NCF)~\cite{he:ncf}.

While iALS scales well with the input data size, it does have a cubic dependency on the embedding dimension $d$, which may make the algorithm infeasible for large $d$.
Follow up work~\cite{pilaszy:als,he:eals,bayer:icd} has addressed this issue and proposed to optimize each coordinate of the embedding vector one at a time which results in a running time with a quadratic dependency on $d$.
Such a coordinate descent algorithm (\ICD) has a clear advantage over \IALS from a complexity point of view.
However, the coordinate-wise updates of \ICD make it hard to take advantage of modern hardware with highly optimized vector processing units, which can benefit \IALS considerably.
Indeed, in Section~\ref{sec:experiments} we empirically investigate this and show that a carefully implemented \IALS algorithm is faster than \ICD unless the dimensionality is very large ($d \approx 1000$).
Up to this point, the advantages of vector processing units outweigh the issues in runtime complexity.
For smaller dimensions $d \approx 100$, \IALS is an order of magnitude faster than \ICD.

In this work, we present a new algorithm, \IALSpp, that combines the benefits of \IALS and \ICD.
\IALSpp optimizes a subvector of the embedding vector at a time.
The size of the subvector can be chosen such that \IALSpp makes use of the vector processing units of modern machines but still preserves a low runtime complexity.
Special cases of the \IALSpp algorithm are scalar optimization, \ICD, where the subvector length is 1, and vanilla \IALS where the subvector length is $d$.

We investigate the properties of \ICD, \IALS and \IALSpp empirically.
We show that \IALSpp shows an order of magnitude faster runtimes than \ICD  for a broad range of embedding dimensions from small $d=64$ to large ones $d=2048$.
The advantages of \IALSpp over traditional \IALS increase as the embedding dimension increases with about an order of magnitude faster speed for $d=800$ and close to two orders of magnitude for $d=2048$.
This indicates that \IALSpp combines a good asymptotic behavior with fast vector processing.
\IALSpp is easy to apply and can be used as a drop in replacement for existing \ICD or \IALS algorithms: neither the final quality nor the convergence are negatively affected; only the runtime improves significantly, as described above.
\IALSpp introduces a new hyperparameter, the subvector length or \emph{block size}, which we show is a purely system related hyperparameter that can be set once and does not need a costly joined search with other learning related parameters.
The best choice of block size is also independent of the embedding dimension -- again this verifies that the block size does not need time consuming tuning with other parameters.

To summarize, we propose \IALSpp, a new algorithm that is better suited for modern vector processing units than \ICD while providing the same improved scaling behavior over traditional \IALS for large embeddings.

\section{Problem Setting}

Our work focuses on item recommendation from implicit feedback where items (e.g., products, movies, songs) from an item catalog $I$ should be recommended to a user $u \in U$.
For learning the recommendation model, there is a historical set of interactions $S \subset U \times I$, where $(u,i) \in S$ indicates that item $i$ has been selected by user $u$ in the past.
To be general, we discuss the setting where every element $(u,i,y,\alpha) \in S$ can have a label $y$ and can be weighted with $\alpha$, i.e., $S \subset U \times I \times \mathbb{R} \times \mathbb{R}^+$.
In most cases, $\alpha=y=1$.
The goal is to learn a scoring function $\hy : U \times I \rightarrow \mathbb{R}$ that assigns a score $\hyu{i}$ to each user-item pair $(u,i)$.
This score can be used for recommending the best $k$ items for a user $u$ by returning the highest scoring items according to $\hy$.

\section{Background}

We start with introducing the matrix factorization model and loss that is used by \IALS.
Then we recap the \IALS~\cite{hu:ials} algorithm and its coordinate descent variation \ICD~\cite{pilaszy:als,he:eals,bayer:icd}.

\subsection{Model and Loss}

The scoring function $\hat{y}$ of matrix factorization is defined as
\begin{align}
   \hyu{i} := \langle \bw_u, \bh_i \rangle, \quad W \in \mathbb{R}^{U \times d}, H \in \mathbb{R}^{I \times d}
\end{align}
Here $d \in \mathbb{N}$ is the embedding dimension, $W$ is the user embedding matrix, $H$ is the item embedding matrix, and $ \langle \bw, \bh\rangle$ denotes the dot product of two vectors $\bw, \bh \in \mathbb R^d$.

The \IALS loss $L(W,H)$ can be defined as:
\begin{align}
    L(W,H) = \sum_{(u,i,y,\alpha) \in S} \alpha (\hyu{i} - y)^2 + \alpha_0\sum_{u \in U} \sum_{i\in I} \hyu{i}^2 + \lambda (\|W\|^2_F + \|H\|^2_F)
\end{align}
The first component of the loss tries to keep the predicted scores of the user-item pairs in the set of observed interactions $S$ close to the observed label $y$.
For implicit feedback problems, where the observed label is usually $y=1$, this loss alone would result in a trivial model, a rank-one solution, that always predicts 1.
However, the items that a user has not rated, i.e., the \emph{unobserved} user-item pairs, are also informative to infer the user's interest.
The second component covers this aspect by trying to keep all\footnote{In the original \IALS formulation, $L_I$ was defined on $(U\times I) \setminus S$ whereas we use all pairs $U \times I$. As shown in~\cite{bayer:icd} both formulations are equivalent when the hyperparameters are adjusted.} predicted scores close to $0$.
Note that this term includes $|U|\cdot|I|$ many pairs.
Finally, a L2 regularizer tries to keep all model parameters close to $0$.
The trade-off between the three components is controlled by the unobserved weight $\alpha_0$ and the regularization weight $\lambda$.

\subsection{Alternating Least Squares Algorithm (\IALS)}
\label{sec:ialsalgorithm}

Hu et al.~\cite{hu:ials} propose to optimize this objective by alternating between optimizing the user embeddings, $W$, and the item embeddings, $H$.
When one side is fixed, e.g., $H$, the problem simplifies to $|U|$ independent linear regression problems, where $\bw_u$ is optimized.
The closed-form solution of each linear regression problem is:
\begin{align}
    \bw_{u^*} \leftarrow \left(\sum_{(u^*,i,y,\alpha) \in S} \alpha \bh_i \otimes \bh_i + \alpha_0 \sum_{i \in I} \bh_i \otimes \bh_i  + \lambda I\right)^{-1} {\sum_{(u^*,i,y,\alpha) \in S} \alpha \bh_i y}. \label{eq:ialsupdate}
\end{align}
A key observation for efficient computation is that the term
\begin{align}
    G^I := \sum_{i \in I} \bh_i \otimes \bh_i,
\end{align}
the Gramian of $H$, is shared between all users and can be precomputed.

Algorithm~\ref{alg:ials} sketches the \IALS algorithm.
The computational complexity to optimize all user embeddings is $\O(d^2\,|S| + d^{3}\,|U|)$.
Learning the item side is analogous to the user side with the same complexity and the overall complexity becomes $\O(d^2\,|S| + d^{3}\,(|U|+|I|))$.
Importantly, the computational complexity is linear in the number of observations $S$, and does not depend on all pairs $|U|\cdot|I|$, which is a key differentiator of \IALS.
Moreover \IALS is trivially parallelizable because of the $|U|$ independent linear regression problems (see line~\ref{ln:ials_user_loop}).
For an efficient distribution over multiple machines, distributed (read only) memory $H$ is required.
Finally, \IALS is a second-order method that converges fast and in typical applications requires only a few alternating steps.  
All these properties make \IALS one of the most scalable and efficient machine learning based collaborative filtering algorithms.

\begin{algorithm}[t]
\caption{\IALS algorithm \cite{hu:ials}\label{alg:ials}}
\begin{algorithmic}[1]
    \For {$t \in \{1,\ldots,T\}$}
        \State $G^I \leftarrow \sum_{i \in I} \bh_i \otimes \bh_i$ \Comment{$\O(|I| d^2)$}
        \For{$u^* \in U$} \label{ln:ials_user_loop}
            \State $\grad{\bw_{u^*}} \leftarrow 0$
            \State $\gradgrad{\bw_{u^*}} \leftarrow \alpha_0  G + \lambda$ \Comment{$\O(d^2)$}
            \For{$(u^*,i,y,\alpha) \in S$}
                \State $\grad{\bw_{u^*}} \leftarrow \grad{\bw_{u^*}}  +  \alpha y \bh_i$ \Comment{$\O(d)$}
                \State $\gradgrad{\bw_{u^*}} \leftarrow \gradgrad{\bw_{u^*}}  +  \alpha \bh_{i} \otimes \bh_{i}$ \Comment{$\O(d^2)$}
            \EndFor
            \State $\bw_{u^*, \bpi} \leftarrow (\gradgrad{\bw_{u^*}})^{-1} \grad{\bw_{u^*}}$\Comment{$\O(d^{3})$}
        \EndFor
        \State{Perform a similar pass over the item side}
    \EndFor
\end{algorithmic}
\end{algorithm}

\subsection{Coordinate Descent Algorithm (\ICD)}

\begin{algorithm}[t]
\caption{\ICD algorithm\label{alg:icd}}
\begin{algorithmic}[1]
    \For {$t \in \{1,\ldots,T\}$}
        \State Compute $\boldsymbol{\hy}$ \Comment{$\O(|S|d)$}
        \For {$f^* \in \{1, \ldots,d\}$}
            \State $\bg^{I} \leftarrow \sum_{i \in I} \bh_{i,f^*} \bh_{i}$ \Comment{$\O(|I| d)$}
            \For{$u^* \in U$}
                \State $\grad{w_{u^*, f^*}} \leftarrow \alpha_0 \langle \bw_{u^*}, \bg^I \rangle + \lambda w_{u^*, f^*}$ \Comment{$\O(d)$}
                \State $\gradgrad{w_{u^*, f^*}} \leftarrow \alpha_0   g_{f^*} + \lambda$ \Comment{$\O(1)$}
                \For{$(u^*,i,y,\alpha) \in S$}
                    \State $\grad{w_{u^*,f^*}} \leftarrow \grad{w_{u^*,f^*}}  +  \alpha (\hy(u^*,i) - y) h_{i,f^*}$ \Comment{$\O(1)$}
                    \State $\gradgrad{w_{u^*,f^*}} \leftarrow \gradgrad{w_{u^*,f^*}}  +  \alpha\, h_{i,f^*}^2 $ \Comment{$\O(1)$}
                \EndFor
                \State $\delta \leftarrow (\gradgrad{w_{u^*, f^*}})^{-1} \grad{w_{u^*, f^*}}$ \Comment{$\O(1)$}
                \State $w_{u^*, f^*} \leftarrow w_{u^*, f^*} - \delta$
                \For{$(u^*,i,y,\alpha) \in S$}
                    \State $\hy(u^*,i) \leftarrow  \hy(u^*,i) - \delta h_{i, f^*}$ \Comment{$\O(1)$}
                \EndFor
            \EndFor
            \State{Perform a similar pass over the item side}
        \EndFor
    \EndFor
\end{algorithmic}
\end{algorithm}

A downside of the original \IALS algorithm is its quadratic/cubic runtime dependency on the embedding dimension, $d$, which becomes an issue for large embedding dimensions~\cite{pilaszy:als}.
As a solution to reduce the runtime complexity, past work~\cite{pilaszy:als,he:eals,bayer:icd} has proposed to apply coordinate descent variations, where a single element, $w_{u^*,f^*}$, of the embedding vector is optimized at a time.
When all parameters but one element, $w_{u^*,f^*}$, are fixed, the optimal value for this coordinate is
\begin{align}
    w_{u^*, f^*}^* = w_{u^*, f^*} - \delta  \label{eq:icd_update}
\end{align}
with
\begin{align}
    \delta = \frac{\sum_{(u^*,i,y,\alpha) \in S} \alpha (\hyu{i} - y) h_{i,f^*} + \alpha_0 \langle \bw_{u^*}, \bg^{I} \rangle  + \lambda w_{u^*, f^*}}{ \sum_{(u^*,i,y,\alpha) \in S} \alpha h_{i,f^*}^2 + \alpha_0 g_{f^*} + \lambda}.
\end{align}
This update rule needs only access to the diagonal $\bg^I \in \mathbb{R}^d$ of the item Gramian $G^I$.
In addition it needs access to the predictions $\hyu{i}$ over observed user-item pairs $(u,i) \in S$.
These can be precomputed and updated with a simple delta update, whenever a coordinate $w_{u^*, f^*}$ changes to $w_{u^*, f^*}^*$:
\begin{align}
    \hy(u^*,i) \leftarrow \hy(u^*,i) + (w_{u^*, f^*}^*-w_{u^*, f^*}) h_{i,f^*}
\end{align}
A degree of freedom when designing coordinate descent algorithms for factorization models is the traversal scheme between dimensions, and when to switch between users and items.
The two main options are:
(i)~alternate between users and items in the outermost loop and iterate over the embedding dimensions in the inner loop, or
(ii)~iterate over embedding dimensions in the outer loop and alternate between users and items in the inner loop.
The second scheme has been found to be better~\cite{rendle:alsfm,yu:ccd} and we will use it in our discussion of \ICD.

The full \ICD algorithm is sketched in Algorithm~\ref{alg:icd}.
Its runtime complexity is $\O(d|S| + d^2\,(|U|+|I|))$ ~\cite{pilaszy:als,he:eals,bayer:icd}.
While this reduces the theoretical complexity compared to \IALS, modern hardware is optimized for vector arithmetic, and scalar treatment has its drawbacks.
We investigate the trade-off between runtime complexity and hardware effects in more detail in Section~\ref{sec:experiments}.

\section{\IALSpp: Subvector Algorithm}
\label{sec:ialspp}

We now present \IALSpp, an algorithm that combines the benefits of vector processing of \IALS and the better runtime complexity for large dimensions of \ICD.
The idea is to optimize subvectors of the embedding vectors at a time.
The size of the subvector is chosen such that the algorithm makes use of the vector processing units.
We first sketch the high level idea and then describe \IALSpp in detail.

For any subvector $\bttheta \subseteq W$ or $\bttheta \subseteq H$, the loss $L$ is quadratic in $\bttheta$, thus the solution to the optimization problem
\begin{align}
    \bttheta^* = \argmin_{\bttheta} L(W,H)
\end{align}
can be obtained by single Newton step
\begin{align}
    \bttheta^* \leftarrow \bttheta - (\gradgradttheta L(W,H))^{-1} \gradbttheta L(W,H) .
\end{align}
For matrix factorization, we can choose $\bttheta$ to be a subvector of the user embedding $\bw_u$ (or the item embedding $\bh_i$).
Let $\bpi \subseteq \{1,\ldots, d\}$ be the indices of this subvector.
We will write $\bw_{u,\bpi}$ to indicate this subvector.
The first and second derivatives of the loss with respect to $\bw_{u^*,\bpi}$ are:
\begin{multline*}
    \frac{1}{2} \grad{\bw_{u^*, \bpi}} L(W,H) = \sum_{(u^*,i,y,\alpha) \in S} \alpha (\hy(u^*,i) - y) \bh_{i,\bpi} \\+ \alpha_0 \bw_{u^*} \sum_{i\in I} \bh_i \otimes \bh_{i,\bpi}  + \lambda \bw_{u^*, \bpi}
\end{multline*}
\begin{align*}
    \frac{1}{2} \gradgrad{\bw_{u^*, \bpi}} L(W,H) &= \sum_{(u^*,i,y,\alpha) \in S} \alpha \bh_{i,\bpi} \otimes \bh_{i,\bpi} + \alpha_0 \sum_{i\in I} \bh_{i,\bpi} \otimes \bh_{i,\bpi} + \lambda .
\end{align*}
Similar to \IALS, we define and precompute (partial) Gramians
\begin{align}
    G^{I,\bpi} := \sum_{i\in I} \bh_i \otimes \bh_{i,\bpi}, \quad G^{I,\bpi} \in \mathbb{R}^{I \times |\bpi|}
\end{align}
in $\O(|I|\,d|\,\bpi|)$ time.
And similar to \ICD, we store and precompute the predictions $\boldsymbol{\hy} \in \mathbb{R}^{|S|}$ in $\O(|S|d)$ time with $\O(|S|)$ space.
Then the sufficient statistics
\begin{multline}
    \frac{1}{2} \grad{\bw_{u^*, \bpi}} L(W,H) = \sum_{(u^*,i,y,\alpha) \in S} \alpha (\hy(u^*,i) - y) \bh_{i,\bpi} \\+ \alpha_0 \bw_{u^*} G^{I, \bpi}  + \lambda \bw_{u^*, \bpi}
\end{multline}
\begin{align}
    \frac{1}{2} \gradgrad{\bw_{u^*, \bpi}} L(W,H) = \sum_{(u^*,i,y,\alpha) \in S} \alpha \bh_{i,\bpi} \otimes \bh_{i,\bpi} + \alpha_0 G^{I, \bpi}_{\bpi} + \lambda
\end{align}
for the solution $\bw_{u, \bpi}^*$ for users $u \in U$ can be computed in $\O(|S| |\bpi| + |U| d |\bpi| + |S| |\bpi|^2)$ time.
Solving the $|U|$ systems of size $|\pi|^2$ takes $\O(|U| |\pi|^{3})$ time.
For computing a new estimate of all model parameters $W,H$, we iterate over both the user and item side and over all subsets of indices (i.e., blocks).
We suggest that the outermost loop iterates over blocks, and within this loop we alternate over user and item side.
This iteration scheme has been found useful for coordinate descent algorithms (e.g., \cite{rendle:alsfm,yu:ccd}).
Algorithm~\ref{alg:ialspp} sketches the full procedure.

\begin{algorithm}[t]
\caption{\IALSpp algorithm\label{alg:ialspp}}
\begin{algorithmic}[1]
    \For {$t \in \{1,\ldots,T\}$}
        \State Compute $\boldsymbol{\hy}$ \Comment{$\O(|S|\,d)$} 
        \For {$\bpi \in P$} \Comment{Iterate over a partition of $\{1,\ldots,d\}$}
            \State $G^{I,\bpi} \leftarrow \sum_{i \in I} \bh_i \otimes \bh_{i,\bpi}$ \Comment{$\O(|I|\,d\, |\bpi|)$}
            \For{$u^* \in U$}
                \State $\grad{\bw_{u^*, \bpi}} \leftarrow \alpha_0 \bw_{u^*} G^{I,\bpi} + \lambda \bw_{u^*, \bpi}$ \Comment{$\O(d\,|\bpi|)$}
                \State $\gradgrad{\bw_{u^*, \bpi}} \leftarrow \alpha_0   G^{I,\bpi}_{\bpi} + \lambda$ \Comment{$\O(|\bpi|^2)$}
                \For{$(u^*,i,y,\alpha) \in S$}
                    \State $\grad{\bw_{u^*, \bpi}} \leftarrow \grad{\bw_{u^*, \bpi}}  + \alpha (\hy(u^*,i) - y) \bh_{i,\bpi}$ \hspace{-0.5mm}\Comment{$\O(|\bpi|)$}
                    \State $\gradgrad{\bw_{u^*, \bpi}} \leftarrow \gradgrad{\bw_{u^*, \bpi}}  +  \alpha \bh_{i,\bpi} \otimes \bh_{i,\bpi}$ \Comment{$\O(|\bpi|^2)$}
                \EndFor
                \State $\bdelta \leftarrow (\gradgrad{\bw_{u^*, \bpi}})^{-1} \grad{\bw_{u^*, \bpi}}$ \Comment{$\O(|\bpi|^{3})$}
                \State $\bw_{u^*, \bpi} \leftarrow \bw_{u^*, \bpi} - \bdelta$
                \For{$(u^*,i,y,\alpha) \in S$}
                    \State $\hy(u^*,i) \leftarrow  \hy(u^*,i) - \langle \bdelta, \bh_{i, \bpi} \rangle$ \Comment{$\O(|\bpi|)$}
                \EndFor
            \EndFor
            \State{Perform a similar pass over the item side}
        \EndFor
    \EndFor
\end{algorithmic}
\end{algorithm}

\begin{table*}[t]
    \centering
    \caption{Summary of the properties of the three algorithms.
    \label{tbl:algorithm_complexity}}
    \setlength{\tabcolsep}{4pt}
    \begin{tabular}{|llll|}
        \hline
         & Runtime Complexity  & Memory & Vector \\
         & per Training Epoch  &  & Instructions \\
        \hline
         \IALS & $ \O(d^2\,|S| + d^3\,(|U|+|I|))$ & $d\,(|U|+|I|)$ & Yes \\
         \ICD & $\O(d|S| + d^2\,(|U|+|I|))$ & $d\,(|U|+|I|) + |S|$  &No \\
         \IALSpp & $ \O\left(d\,|\bpi|\,|S| + (d^2  + d|\bpi|^2)\,(|U|+|I|) \right)$ & $d\,(|U|+|I|) + |S|$  & Yes \\
        \hline
    \end{tabular}
\end{table*}

The overall computational complexity per epoch is:
\begin{align}
    &\O\left(|S| d + \frac{d}{|\bpi|}\left((|U|+|I|) (d |\bpi| + |\bpi|^2 + |\bpi|^{3}) + |S|(|\bpi|+ |\bpi|^2)\right)\right) \notag \\
    \equiv &\O\left(d|S||\bpi| + (d^2  + d|\bpi|^{2})\,(|U|+|I|)  \right)
\end{align}
Notable special cases are 
    \ICD, $|\bpi| = 1$, with $\O(d|S| + (|U|+|I|)d^2)$~\cite{pilaszy:als,he:eals} 
    and
    vanilla \IALS, $|\bpi| = d$, with $\O(d^2|S| + (|U|+|I|)d^{3})$~\cite{hu:ials}.
Table~\ref{tbl:algorithm_complexity} summarizes the algorithms.
From a pure complexity point of view and ignoring convergence speed, $|\bpi|=1$  and more generally any $|\bpi|=\text{const}$ are in the best complexity class.
On modern hardware the cost for solving one block of $|\bpi|=p$ might be much faster than solving $p$ blocks of $|\bpi|=1$, so a (constant) block size larger than 1 can be beneficial.
Complexity per step is not the only factor but speed of convergence needs to be considered as well.
Vanilla \IALS with $|\bpi|=d$ directly calculates the optimum for a side $W$ (or $H$), but for smaller blocks, $|\bpi|<d$, the optimum is only within the block.
If we would iterate over all the blocks in $W$ before switching to optimizing the blocks in $H$, then the larger block sizes would have faster convergence.
However, the scheme we implement (i.e. optimizing over $W$ and $H$ in the same block before switching to another block) can counter the downside of using smaller blocks.
We evaluate the effect of the block size on convergence in Section~\ref{sec:experiments_convergence}.

\paragraph{Relationship To Block Coordinate Descent}

The alternating least squares algorithms that are discussed in this paper, \ICD, \IALS, \IALSpp are instances of block coordinate descent.
Block coordinate descent~\cite{tseng:bcd} is a class of optimization algorithms that updates a subset of the model parameters while keeping the remaining parameters fixed.
For matrix factorization, block coordinate descent is not limited to the three choices of subsets (scalar $w_{u}$, embedding vector $\bw_{u}$, subset of an embedding vector $\bw_{u,\bpi}$) that we discuss in this paper but would include any subset of $W,H$ potentially even including subsets that overlap both $W$ and $H$.
Block coordinate descent has also been studied in the context of non-negative matrix factorization.
For example \cite{kim:nnmfbcd} discusses three choices for blocks: scalars (as in \ICD), embedding vectors (as in \IALS) and whole matrices $W$ (or $H$).
A general template for deriving implicit alternating least squares algorithms for a broader class of recommender models has been discussed in~\cite{rendle:implicit}.
The \IALSpp algorithm presented in this paper is a concrete instance of this general algorithm where a subvector of a user or item embedding is chosen as the block.

\section{Evaluation}
\label{sec:experiments}

We empirically study the properties of the different algorithms \IALS, \ICD and \IALSpp as well as the impact of the block size, $\bpi$.
We are particularly interested in the following effects:
\begin{enumerate}
    \item \textbf{Wall runtime for one training epoch}:
    The theoretical complexity analysis suggests that \ICD is faster than \IALS and is in the same class as \IALSpp.
    We want to provide wall time measurements with a real implementation on a real dataset.
    \item \textbf{Final quality}:
    The three algorithms are non-convex and it is unclear if they can achieve comparable quality.
    In this analysis, we will discuss quality independent of runtime.
    \item \textbf{Convergence speed}:
    Even if the same final quality can be achieved, the convergence speed might differ.
    We provide results both for convergence speed in number of epochs and for wall runtime.
\end{enumerate}

\subsection{Setup}

\subsubsection{Dataset and Evaluation Protocol}

\begin{table*}[t]
    \centering
    \caption{Benchmarks (dataset and evaluation protocol) used in our experiments. Hyperparameters are from \citetuning{}.
    \label{tbl:data}}
    \setlength{\tabcolsep}{4pt}
    \begin{tabular}{|lc|rrr|SSr|}
    \hline
        \multicolumn{2}{|c|}{Benchmark} & \multicolumn{3}{c|}{Data Statistics}& \multicolumn{3}{c|}{Hyperparameters} \\
        Dataset       & Eval from & $|U|$  & $|I|$ & $|S|$ & {$\lambda$} & {$\alpha_0$} & T \\
        \hline
        ML20M~\cite{harper:movielens}              & \cite{liang:vae} & 136,677 & 20,108 & 10.0M & 0.003 & 0.1  & 16 \\
        MSD\cite{bertinmahieux:msd}                & \cite{liang:vae} & 571,355 & 41,140 & 33.6M & 0.002 & 0.02 & 16\\
        \hline
    \end{tabular}
\end{table*}

We use the ML20M~\cite{harper:movielens} and MSD~\cite{bertinmahieux:msd} datasets with the experimental protocol from~\cite{liang:vae} (see Table~\ref{tbl:data} for dataset statistics).
The split and protocol of~\cite{liang:vae} holds out a set of users for tuning and testing purposes.
At evaluation time, the recommender is given 80\% of the interactions of each of the holdout users and is asked to generate recommendations.
For each user, the position of the remaining 20\% items in the predicted ranked list is measured and Recall and NDCG metrics are computed.
This benchmark has been used by several authors and we follow the protocol strictly, so our quality results can be compared to other publications.
For more details on the evaluation protocol, see~\cite{liang:vae} and their published evaluation code.

We use the \IALS hyperparameters from~\citetuning{}:
See Table~\ref{tbl:data} for values of regularization $\lambda$ and unobserved weight $\alpha_0$.
We train for $T=16$ epochs, and use an initialization of $\sigma^*=0.1$ and regularization scaling $\nu=1$ as explained in~\citetuning{}.
We study the effect of the embedding dimension $d \in \{64,128,256,512,1024,2048\}$ and block size $|\bpi|=\{1, 2, 4, 8, 16,32, \ldots, d\}$.
To reduce noise for the ML20M results, we repeat all experiments 4 times and report averages.
The MSD benchmark has 5 times more training and holdout users and thus shows less noise, so we did not repeat the experiments for MSD multiple times.

\subsubsection{Comparison to Other Collaborative Filtering Methods}

The focus of our work is not a qualitative comparison to other collaborative filtering methods but to compare different optimization methods for \IALS.
We refer the interested reader to~\citetuning for detailed comparisons of \IALS to other collaborative filtering methods including different variational autoencoders~\cite{liang:vae}, EASE~\cite{steck:ease}, NCF~\cite{he:ncf} and other optimization methods for matrix factorization such as LambdaNet~\cite{burges:lambdanet} or WARP~\cite{weston:wsabie}.
The study in \citetuning{} shows that \IALS achieves competitive quality compared to these other collaborative filtering methods.
The quality results for \IALS that we obtain in the experiments here match the ones provided in~\citetuning{}, so all the findings with respect to other collaborative filtering methods presented in~\citetuning{} also apply to this work.

\subsubsection{Implementation}

For any wall time study, the implementation plays a major role.
We implement the algorithms for \IALS (Algorithm~\ref{alg:ials}), \ICD (Algorithm~\ref{alg:icd}) and \IALSpp (Algorithm~\ref{alg:ialspp}) in C++\footnote{Source code available at: \url{\sourcecodelink}}.
The implementations are single machine but multithreaded and use Eigen\footnote{\url{https://eigen.tuxfamily.org/}} for vector and matrix operations supporting AVX instructions.
Note that while \IALSpp subsumes the other two algorithms, i.e., \IALS with $d=|\bpi|$ and \ICD with $d=1$, the numbers that we report for \IALS and \ICD are for customized and further optimized implementations.
For example, for \IALS, the computation of $\boldsymbol{\hy}$ can be skipped and for \ICD we found it beneficial to hard code the scalar implementation for all operations instead of relying on vector operations of length 1.
We observe that these further optimizations have a positive effect on the runtime compared to using \IALSpp with $d=1$ or $d=|\bpi|$.
All time measurements reported in this paper were obtained on the same machine, a standard desktop computer.
We use only the CPU and don't use any GPU.

\subsection{Results}

\subsubsection{Wall time per Epoch}

\begin{figure*}[t]
    \centering
    \subfigure[Embedding Dimension $d$]{\includegraphics[width=\myfigurewidth]{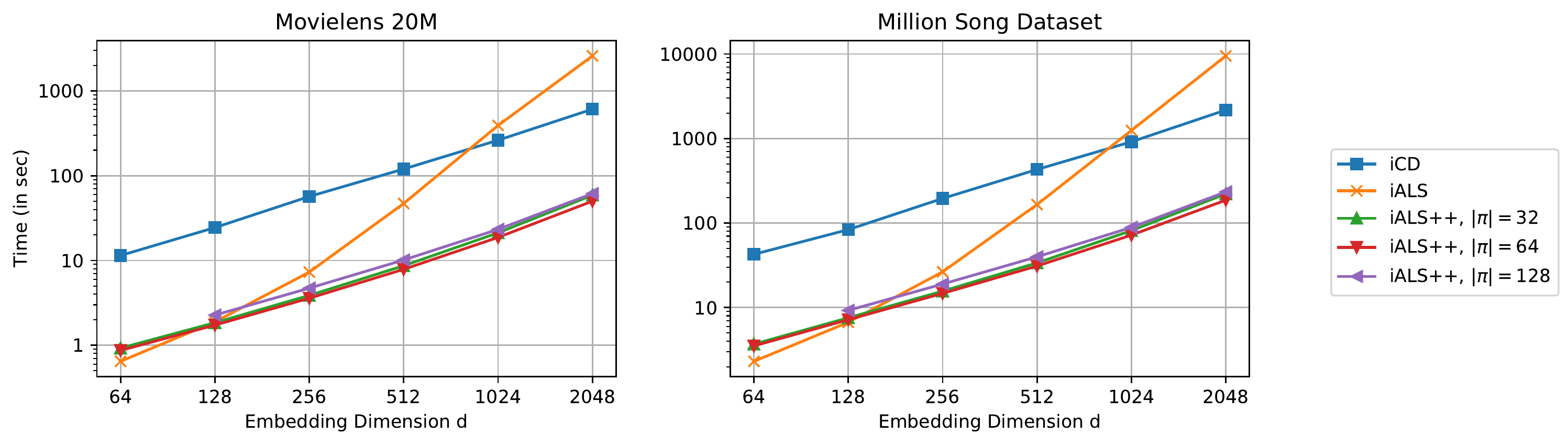}}\\
    \subfigure[Block Size $|\bpi|$]{\includegraphics[width=\myfigurewidth]{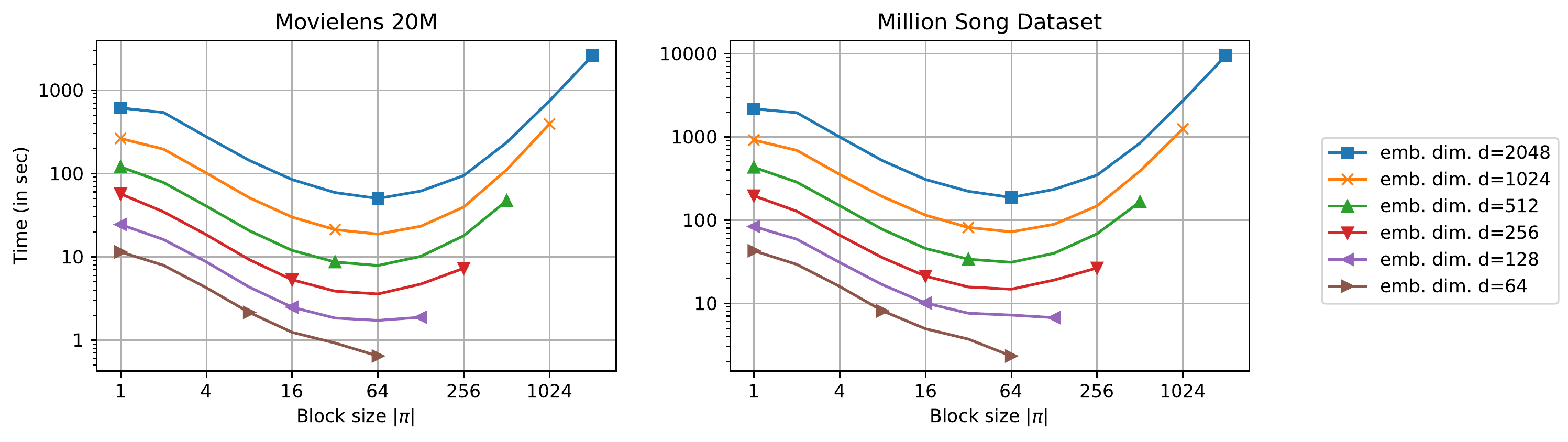}}%
    \caption{Runtime comparison of \ICD, \IALS and \IALSpp (varying the block size $|\bpi|$) for different embedding dimensions, $d$.
    \IALSpp with a moderate block size (e.g., $|\bpi|=64$) is about one order of magnitude faster than \ICD.
    The advantages of \IALSpp over \IALS increase with the embedding dimension $d$.
    Note that the axes are logarithmic.}
    \label{fig:block_size_vs_time}
\end{figure*}

The theoretical complexity analysis in Table~\ref{tbl:algorithm_complexity} provides the asymptotic runtime dependency of the three algorithms \ICD, \IALS and \IALSpp with respect to the embedding dimension $d$.
As in any asymptotic analysis, the complexity results give an indication of the behavior in the limit.
It does not imply that an algorithm with a better complexity class is always the better choice.
Next, we investigate the behavior of real implementations on real data.

Figure~\ref{fig:block_size_vs_time} shows runtime results that we obtained for the three algorithms on the Movielens 20M and Millions Song benchmarks.
The top plots (Figure~\ref{fig:block_size_vs_time}(a)) show the runtime for one full epoch, i.e., one iteration over the outermost loop of Algorithms~\ref{alg:icd}, \ref{alg:ials} and \ref{alg:ialspp}.
For smaller embedding dimensions ($d<256$), \IALS is about an order of magnitude faster than \ICD.
This difference can be explained by \IALS being better suited for vector processing units.
For larger dimensions, the advantages disappear, and \ICD becomes faster than \IALS at $d\approx800$.
The empirical results of \IALS and \ICD are in line with the asymptotic analysis which states that for large enough dimensions, \ICD will have a better performance than \IALS.
However, it also reveals that the better asymptotic behavior of \ICD comes at a high costs for smaller dimensions, where \IALS is clearly a better choice.
\IALSpp combines the advantages of \ICD and \IALS and is about an order of magnitude faster than \ICD throughout all embedding dimensions.
This is consistent with the complexity analysis where \ICD and \IALSpp both have the same dependency on the embedding dimension.
\IALSpp also shows a fast runtime for smaller dimensions comparable to \IALS.

The bottom two plots of Figure~\ref{fig:block_size_vs_time} show results for all combinations of block sizes and embedding dimensions\footnote{The results for $|\bpi|=1$ were obtained with an \ICD solver and the results for $|\bpi|=d$ with an \IALS solver.}.
The U-shape of each curve verifies that there is a trade-off between runtime complexity (\ICD, $|\bpi|=1$) and vector operations (\IALS, $|\bpi|=d$).
The best choice is achieved by controlling this trade-off with the block size that \IALSpp introduces.
For all curves, a medium sized block size of around 64 works best.
This indicates that the best choice for the block size is a hardware dependent constant.

\subsubsection{Final Quality}

\begin{figure*}[t]
    \centering
    \includegraphics[width=\myfigurewidth]{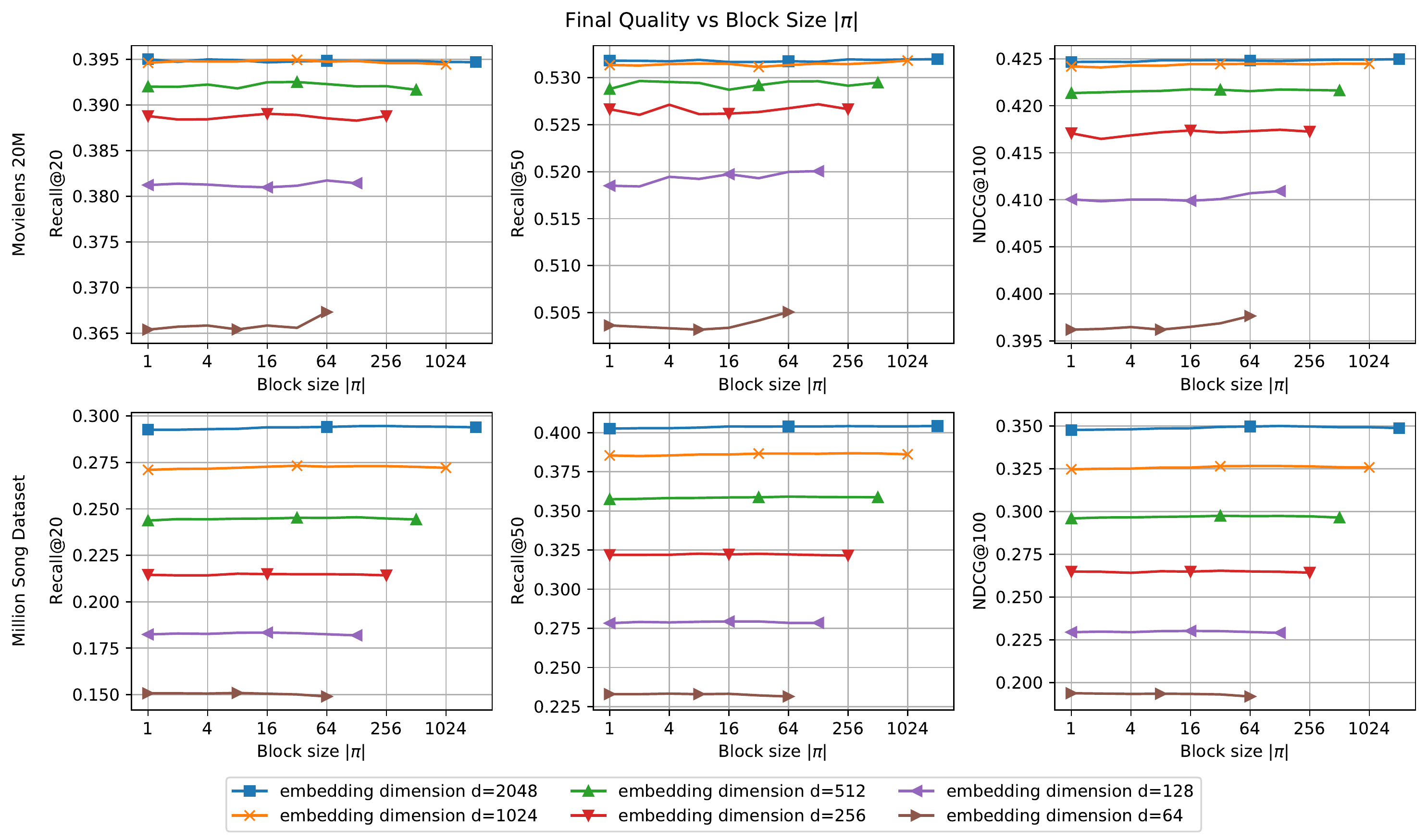}%
    \caption{The three different algorithms (\IALS, \ICD, \IALSpp) and the block size $|\bpi|$ have no noticeable effect on the final quality of the model.
    Note that the results for block size $|\bpi|=1$ were generated with \ICD and the ones for $|\bpi|=d$ with \IALS, all other results are using \IALSpp.}
    \label{fig:block_size_vs_quality}
\end{figure*}

The discussion so far focused on the runtime of the algorithms but ignored the quality.
The algorithms discussed in this work, \ICD, \IALS and \IALSpp, solve non-convex problems and our analysis does not provide guarantees on the final quality of these algorithms nor on the convergence speed.
We examine both problems empirically.
Figure~\ref{fig:block_size_vs_quality} shows the final quality for all embedding sizes with a varying block size.
The results for $|\bpi|=1$ where obtained with \ICD, the results for $|\bpi|=d$ with \IALS and all the remaining ones with \IALSpp.
As can be seen, the curves are mostly flat which indicates that all three optimization algorithms can produce models with the same quality.
That means the solver algorithm can be selected based on the training speed.

\subsubsection{Convergence Speed}
\label{sec:experiments_convergence}

\begin{figure*}[t]
    \centering
    \includegraphics[width=\myfigurewidth]{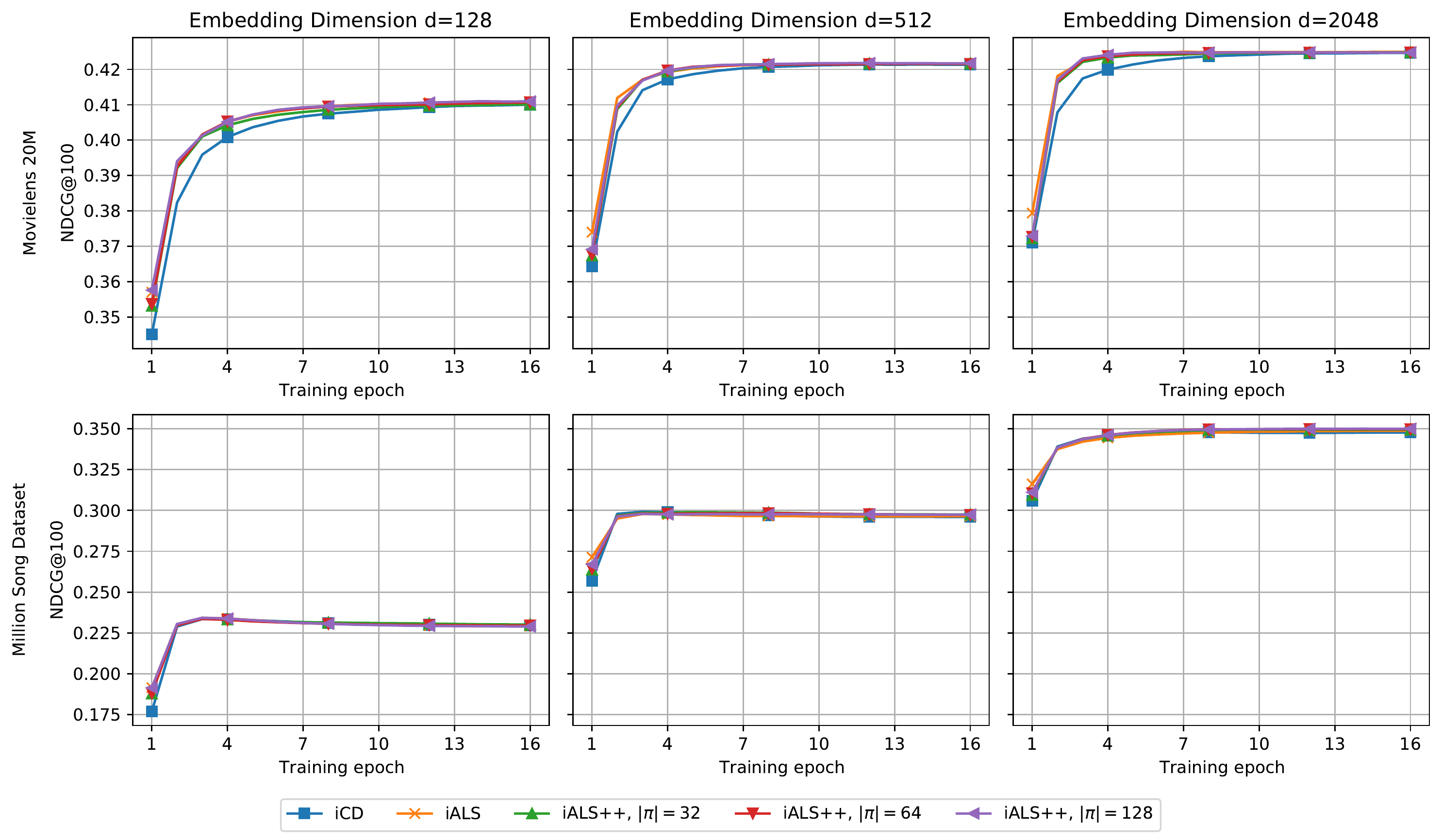}%
    \caption{The three different algorithms (\IALS, \ICD, \IALSpp) and the block size $|\bpi|$ have no noticeable effect on convergence behavior with respect to training epochs.
    However the time per epoch reduces with a good block size (see Figure~\ref{fig:block_size_vs_time}) making the convergence in wall time much faster (see Figure~\ref{fig:convergence_walltime}).}
    \label{fig:convergence_epochs}
\end{figure*}

So far, we have covered the runtime for one epoch and the final quality.
Now, we examine how fast the algorithms converge to their solutions.
Figure~\ref{fig:convergence_epochs} shows the progress of the prediction quality after each training epoch.
We show the results for the NDCG@100 measure and for three embedding dimensions (small $d=128$, medium $d=512$ and large $d=2048$).
We omit the convergence results for Recall@20 and Recall@50 due to space constraints, but all curves look very similar.
As can be seen, the three algorithms have a very similar convergence behavior and are well converged within 16 epochs.
\ICD shows a slightly slower convergence than \IALS and \IALSpp with a block size of about $64$.
It is important to note that this plot ignores that each algorithm has vastly different runtimes per epoch (see Figure~\ref{fig:block_size_vs_time}).
The purpose of the plot is to give implementation independent insights into the convergence behavior.

\begin{figure*}[t]
    \centering
    \includegraphics[width=\myfigurewidth]{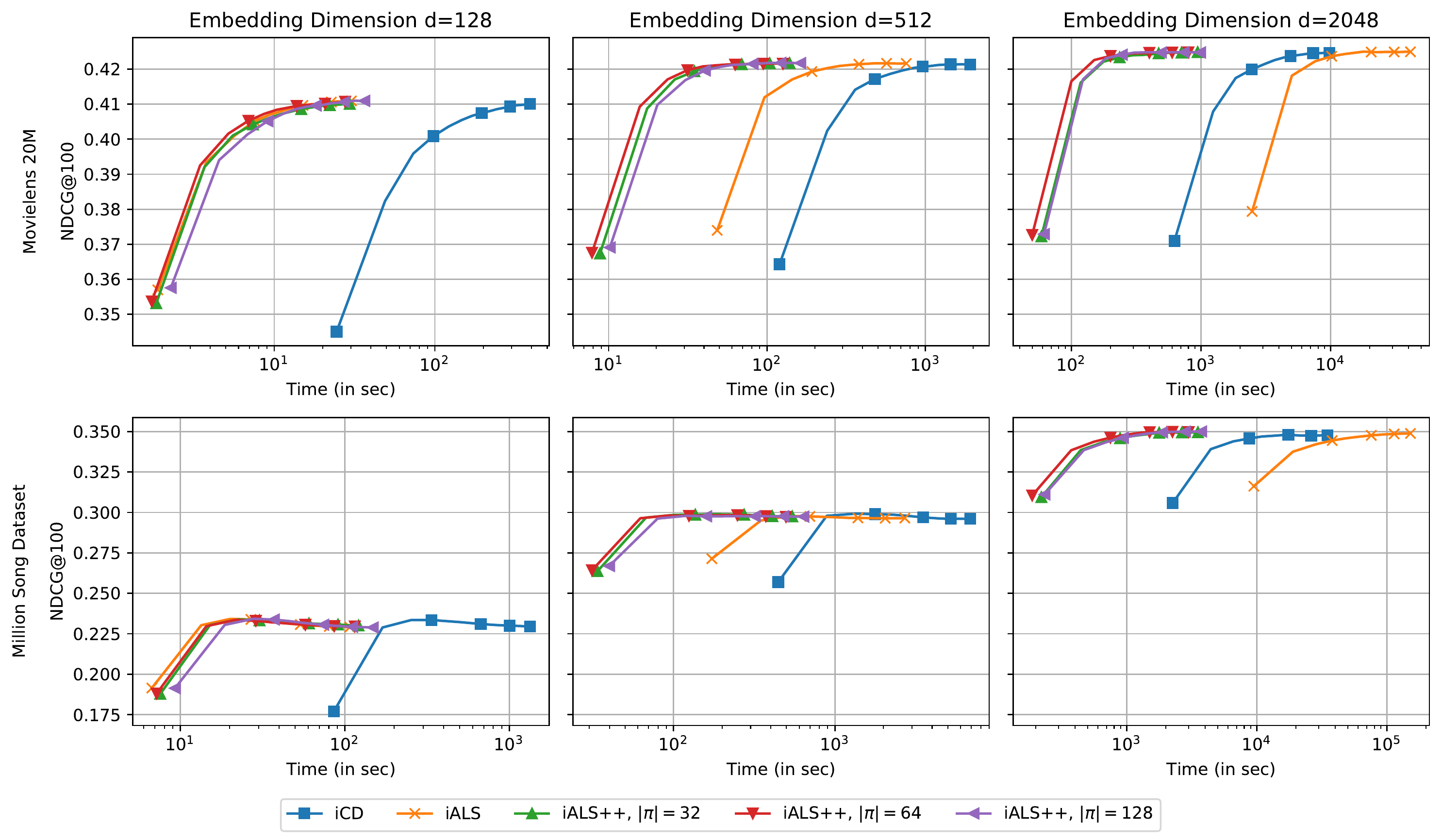}%
    \caption{\IALSpp converges in an order of magnitude less time than \ICD. The advantages of \IALSpp over \IALS increase with the embedding dimension $d$.
    }
    \label{fig:convergence_walltime}
\end{figure*}

Finally, Figure~\ref{fig:convergence_walltime} shows the convergence curves with respect to wall time.
In practice, this is the measurement that matters most.
As expected from our results so far, the faster runtime per epoch of \IALSpp leads to a much faster overall convergence.
For small dimensions ($d=128$), \IALSpp and \IALS are an order of magnitude faster than \ICD.
That means when \ICD has finished just one epoch, \IALSpp and \IALS have already converged.
For mid sized dimensions ($d=512$), \IALS starts to suffer from the higher runtime complexity and the advantages over \ICD slowly vanish.
In contrast to this, \IALSpp is still about an order of magnitude faster than \ICD and much faster than \IALS.
For large embedding dimensions ($d=2048$), \IALSpp is one order of magnitude faster than \ICD and nearly two orders of magnitude faster than \IALS.

The experiments also give further evidence that ALS type solvers are very efficient for learning recommender systems from implicit feedback.
A fully converged model for the ML20M dataset with 20k items and 136k users can be learned in less than a minute for $d=128$, about 3 minutes for $d=512$ and about 15 minutes for $d=2048$.
These numbers are achieved with a single machine implementation without GPU or other non-CPU accelerators.

\subsection{Comparison to Previous Studies on \ICD and \IALS}

Our experiments also give new insights into the behavior of the existing algorithms \ICD and \IALS.
Previous studies \cite{pilaszy:als,he:eals} have concluded that \ICD is a much faster algorithm for all embedding dimensions than \IALS.
Our study with implementations that are optimized for vector processing units has different findings.
For example, \cite{pilaszy:als} studies a problem of size $|S|=9,617,414$ and $|U|+|I| = 289,493$, and \cite{he:eals} studies a problem of size $|S|=5,020,705$ and $|U|+|I|=192,565$.
These sizes are comparable to the ML20M dataset -- the differences are up to a two times smaller number of user-item pairs $|S|$ (for \cite{he:eals}) and up to two times larger number of users and items $|U|+|I|$ (for \cite{pilaszy:als}).
The datasets in \cite{pilaszy:als,he:eals} are smaller than MSD in both dimensions.
For $d \approx 100$, both studies show that \ICD is about 10x faster than \IALS:
\cite{pilaszy:als} reports 68 sec for one epoch for \ICD (``IALS1'' in Table~2 of \cite{pilaszy:als}) and 572 sec for \IALS.
\cite{he:eals} reports 72 sec for one epoch of \ICD (``eALS'' in Table~3 of \cite{he:eals}) and 1260 sec for \IALS.
Our results for $d=128$ and ML20M are 24 sec for one epoch of \ICD and 2 sec for one epoch of \IALS.
While the numbers for the \ICD results roughly match, our results for \IALS are more than 100x faster.
We suspect that the large differences in \IALS results are due to missing vector optimization in previous implementations that are now available in modern hardware and numerical libraries.
If an implementation does not make use of vector processing units then the observed runtime is likely closer to the theoretical O-notation results as previously observed.

Finally, ignoring runtime per epoch, the convergence behavior that we observe in Section~\ref{sec:experiments_convergence} matches the findings of \cite{pilaszy:als}: both \IALS and \ICD show more or less the same convergence speed with very slight advantages for \IALS and the final quality is mostly indistinguishable.

\section{Conclusion}

In this work we showed the benefits of utilizing vector processing when optimizing the \IALS loss.
Unlike \ICD, the traditional \IALS algorithms is well suited for vector processing and shows an order of magnitude faster runtime than \ICD for small size embedding dimensions ($d \approx 100$).
However, this advantage vanishes for larger embeddings (consistent with its theoretical runtime complexity) and \IALS becomes very slow for large embedding dimensions ($d \gg 1000$).

We propose a new algorithm \IALSpp that combines the advantages of vector arithmetic of \IALS with a low runtime complexity as in \ICD.
In our experiments, \IALSpp shows an order of magnitude faster convergence speed than \ICD both over small and large embedding dimensions.
For small embedding dimensions, \IALSpp is as fast as \IALS but as expected from the theory, has a large advantage over \IALS as the embedding dimension increases.

Our experiments indicate that the algorithms do not have undesirable side effects on final quality or per epoch convergence.
Instead, the algorithm can be solely picked based on speed considerations.
Throughout the experiments, \IALSpp is much faster than the alternatives, \IALS and \ICD.
Unless an application is interested only in small embedding dimensions ($d \approx 100$) where \IALS performs as well as \IALSpp, \IALSpp is clearly the better choice.
For implementations that are optimized for vector processing, \ICD is a poor choice and \IALS and \IALSpp should be preferred.

Finally, in recent years, stochastic gradient descent (SGD) based machine learning frameworks such as TensorFlow have been heavily optimized for vector processing hardware.
SGD is well suited to training models with large embedding dimensions as it does not require a costly matrix solve.
However, it suffers from slower convergence rates compared to ALS based methods.
The classic \IALS solver is well suited to vector processing but has a problematic computational complexity and becomes prohibitively slow when the embedding dimension is large.
Our proposed \IALSpp variation brings the benefits of vector processing to large embedding models, while maintaining the advantages of second order optimization.

\bibliographystyle{acm}

\end{document}